\documentclass{article}

\usepackage{arxiv}

\usepackage[utf8]{inputenc} 
\usepackage[T1]{fontenc}    
\usepackage{hyperref}       
\usepackage{url}            
\usepackage{booktabs}       
\usepackage{amsfonts}       
\usepackage{nicefrac}       
\usepackage{microtype}      
\usepackage{lipsum}		
\usepackage{graphicx}
\usepackage{doi}
\usepackage{multirow}
\usepackage{float}
\usepackage{algorithm,algpseudocode}

\title{Explainable expected goal models for performance analysis in football analytics}

\author{ \href{https://orcid.org/0000-0002-6172-5449}{\includegraphics[scale=0.06]{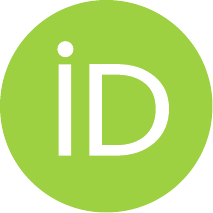}\hspace{1mm}Mustafa~Cavus}\thanks{The work on this paper is financially supported by the NCN Sonata Bis-9 grant 2019/34/E/ST6/00052} \\
	Warsaw University of Technology\\
	Faculty of Mathematics and Information Science\\\vspace{2mm}
	Warsaw, Poland\\
	Eskisehir Technical University\\
	Department of Statistics\\
	Eskisehir, Turkey\\
	\texttt{mustafacavus@eskisehir.edu.tr} \\
	\\
	\And
	\href{https://orcid.org/0000-0001-8423-1823}{\includegraphics[scale=0.06]{orcid.pdf}\hspace{1mm}Przemysław~Biecek} \\
	Warsaw University of Technology\\
	Faculty of Mathematics and Information Science\\
	Warsaw, Poland\\
	\texttt{przemyslaw.biecek@pw.edu.pl} \\
}

\renewcommand{\shorttitle}{Explainable expected goal models for performance analysis in football analytics}

\hypersetup{
pdftitle={A template for the arxiv style},
pdfsubject={q-bio.NC, q-bio.QM},
pdfauthor={David S.~Hippocampus, Elias D.~Striatum},
pdfkeywords={First keyword, Second keyword, More},
}

\begin{document}
\maketitle

\begin{abstract}
	The expected goal provides a more representative measure of the team and player performance which also suit the low-scoring nature of football instead of score in modern football. The score of a match involves randomness and often may not represent the performance of the teams and players, therefore it has been popular to use the alternative statistics in recent years such as shots on target, ball possessions, and drills. To measure the probability of a shot being a goal by the expected goal, several features are used to train an expected goal model which is based on the event and tracking football data. The selection of these features, the size and date of the data, and the model which are used as the parameters that may affect the performance of the model. Using black-box machine learning models for increasing the predictive performance of the model decreases its interpretability that causes the loss of information that can be gathered from the model. This paper proposes an accurate expected goal model trained consisting of 315,430 shots from seven seasons between 2014-15 and 2020-21 of the top-five European football leagues. Moreover, this model is explained by using explainable artificial intelligence tool to obtain an explainable expected goal model for evaluating a team or player performance. To the best of our knowledge, this is the first paper that demonstrates a practical application of an explainable artificial intelligence tool aggregated profiles to explain a group of observations on an accurate expected goal model for monitoring the team and player performance. Moreover, these methods can be generalized to other sports branches.
\end{abstract}


\section{Introduction}
In recent years, the exponential speed of improvement in the technologies supporting the collection, storage, and analysis of data has a revolutionary effect on football analytics as well as many fields. The easy accessibility of data provides a great potential to propose several key performance metrics measuring several aspects of the play such as pass evaluation, quantifying controlled space, evaluating shots, and goal-scoring opportunities through possession values. One of these prominent metrics is \emph{expected goal} (xG) which is the most notable one in football talkshows in TV and end-of-match statistics nowadays. It is proposed by Green \cite{green} to quantify the probability of a shot being the goal. The reason behind the development of such a metric is to propose a metric to represent the low-scoring nature of football rather than the other sports. It is an ordinary story in football that a team dominated the game -they had many scoring opportunities- but could not goal, and the opponent won the match by converting one of the few goal opportunities to goal they created. In this case, the xG is used as a useful indicator of the score. It can be defined as the mean of a large number of independent observations of a random variable which is the shots from the statistical point of view. Besides being a good representative of the score, it is also a good indicator which is used to predict the future team performance \cite{cardoso}. There are many studies is conducted to train a machine learning model learned from the predictors such as shot type, distance to goal, angle to goal for predicting the value of xG \cite{rathke, tippana, pardo, herbinet, wheatcroft, bransen, sarkar_and_kamath, Eggels}. The data used to train the xG model is usually highly unbalanced. It causes an important problem that seen in some of these papers is the poor prediction performance of the models on minority class \cite{Anzer_and_Bauer}. In this paper, we aimed to propose an accurate xG model in terms of both majority and minority classes.

One of the practical applications of the xG models, which is the main focus of this paper, is performance evaluation \cite{kharrat, spearman}. Brechot and Flepp \cite{brechot} proposed to use the xG models for performance evaluation instead of match outcomes which may easily be influenced by randomness in short-term results. They introduced a chart built upon the concept of the xG by plotting the teams' ranking in the league table against their rankings based on xG. Moreover, they proposed some useful metrics calculated based on xG such as offensive and defensive ratios. Fairchild et al. \cite{fairchild} focused on ways for evaluating the xG model goes beyond the accuracy, which is second from a player and team evaluation perspective on offensive and defensive efficiency by comparing the xG metric with the actual goals. They created the xG model for Major League Soccer in the USA and Canada. However, these papers consider only the output of the xG model. Thanks to the XAI tools, it is possible to explain a black-box machine learning model's behavior at the local and global levels. In this way, we can gather more information from the model not only its prediction and also its behavior. The one of most commonly used tool at local-level is the ceteris-paribus (CP) profiles that show the change of model prediction would change for the value of a feature on a single observation \cite{ema}. Its use to explain only one observation is a limitation, but by aggregating these profiles, it is possible to explain more than one observation at the same time. Actually, partial dependence profile (PDP) is used to explain the relationship between a feature and the response \cite{friedman}. However, the PDP is the estimation of the mean of the CP profiles for all observations in a dataset, not just some of the observations. In football, offensive performance can be measured through the shots taken by a team or player. In this paper, we introduced a practical application of an XAI tool based on the aggregation of the CP profiles  which are used for the local-level explanation of the model behavior. By this approach, we can evaluate a player or team's performance and answer the what-if type questions about the performance. 

The main contributions of this paper are: (1) proposing the most accurate xG model, in terms of both majority and minority classes, trained on the data consists 315,430 shots from seven seasons between 2014-15 and 2020-21 of the top-five European football leagues, and (2) introducing a novel team/player performance evaluation approach which is a practical application of the XAI tools in football based on the aggregation of the CP profiles. We believe that the approaches given in this paper can be generalized for the other branches of sport.

The remainder of this paper is structured as follows: Sec. II introduces the mathematical background of the xG models and xG model training. The performance of the trained xG models are investigated in Sec. III. Lastly, in Sec. IV, we introduce how the aggregated profiles are created and used to evaluate the performance of a player or team. Moreover, we demonstrate a practical application of the aggregated profiles based on the xG model for player and team levels. 

\section{Expected goal models}

Consider $\mathbf{X} \subseteq \mathbb{R}^d$ is a $d$-dimensional feature vector, $Y \in \{0, 1\}$ is the label vector of response variable. The dataset is denoted by $D = \{ (\mathbf{x}_i, y_i) \}^n_{i=1}$ where each sample $(x_i, y_i)$ is independently sampled from the joint distribution with density $p(\mathbf{x}, y)$ which includes an instance $\mathbf{x}_i \in \mathbf{X}$ and a label $y_i \in Y$. The goal of a binary classifier is to train an optimal mapping function as follows:

\begin{equation}
    f: \mathbf{X} \rightarrow Y
\end{equation}

\noindent by minimizing a loss function is $L(f) = P[Y \neq f(\mathbf{X})]$. The xG model is a special case of supervised classification task which has a binary outcome that takes the values are goal or not. Here $Y$ is the target variable which shows the goal or not of a shot, and the $\textbf{X}$ are the features that are used to predict the value of $Y$. The most commonly used features are distance to goal, angle to goal, shot type, last action, etc. The calculation of the xG value from the xG model can be easily algorithmized: (1) the individual scoring probabilities of the shots are calculated, (2) these probabilities are summed over for a player, or a team to derive the cumulative xG value \cite{brechot}. The calculation of xG value for a player/team can be also seen in Algorithm~\ref{alg:xg}. Assume that there is a player or a team which is xG value calculated for. It can be calculated for a game, a season that includes multiple games, or a specific part of the season, e.g. the period after a player recovers from injury. Let $n_i$ is the total number of shots of the player/team in the interested time period, and the features used to train the model is $\textbf{X}_i$ and the response variable is $y$. The predicted values of the model are summed for each shot to calculate the xG of the shots taken by the player/team in a time period.

\begin{algorithm}
\caption{Calculation of the xG value for a player / team}\label{alg:xg}
\begin{algorithmic}[1]
\State \textbf{Input:} $\textbf{X}_i$, $y$, a player / team. 

\State Train an xG model: $y \sim f(\textbf{X}_i)$.

     \For{$i \gets 1$ to $n_i$}
        \State Predict $f(\textbf{X}_i)$ for $i = 1, 2, ..., n_i$
    \EndFor

\State $xG_{player/team} = \sum_{i=1}^{n_i} f(\textbf{X}_i)$
\end{algorithmic}
\end{algorithm}

For example, assuming that a team had three shots in a match with probabilities of 0.50, 0.20, and 0.05 means that the team has generated chances worth 0.75 xG. It can be calculated not only for a match, but also for different time periods such as season(s). This creates many different practical applications' opportunities.

In the following subsections, we describe the data we used to train xG models, and give the steps about the pre-processing of data. Then,  we introduce the tools we used in model training and explanation. Lastly, the problem in the xG model is imbalanced data is discussed and the solution way is mentioned.

\subsection{Description of the Data}

The issue that needs to be discussed, before the training of a xG model, is the characteristics of the data used to train the xG models. It is expected that the style of play changes over time and varies between leagues from the football enthusiasts' point of view. The answer of "How it can be determined whether this situation has occurred or not?" can be found in Robberechts and Davis \cite{Rob_and_Davis}. They conducted an extensive experimental study to investigate the frequently asked data-related questions such as "How much data is needed to train an accurate xG model?", "Are xG models league-specific?", and "Does data go out of date?" that may affect the performance of an xG model. Their results show that five seasons of data are needed to train a complex xG model, the data does not go out of date, and using league-based xG models does not increase the accuracy significantly. We determined our model development strategy considering these findings in this paper. 

We focus in our paper on 315,430 shots-related event data (containing 33,656 goals $\sim 10.66\%$ of total shots) from the 12,655 matches in 7 seasons between 2014-15 and 2020-21 from the top-five European football leagues which are Serie A, Bundesliga, La Liga, English Premier League, Ligue 1. The dataset is collected from Understat\footnote{\href{https://understat.com}{https://understat.com}} by using the R-package worldfootballR \cite{worldfootballR} and excluded the 1,012 shots resulting in own goals due to their unrelated pattern from the concept of the model. The package provides useful functions to gather and handle the shots data by matches, seasons, and leagues from the various data sources such as FBref\footnote{\href{https://fbref.com/en/}{https://fbref.com/en/}}, Transfermarkt\footnote{\href{https://www.transfermarkt.com}{https://www.transfermarkt.com}}, and Fotmob\footnote{\href{https://www.fotmob.com}{https://www.fotmob.com}}. The detailed information, such as type and description, about the features used in the model, are given in Table \ref{tab:features}.

\begin{table}[h]
    \caption{Details of the variables used to train our xG model}
    \begin{center}
    \begin{tabular}{p{3cm} p{2cm} p{9cm}}\toprule
        \textbf{Features} & \textbf{Type} & \textbf{Description} \\\toprule
        \texttt{status}            & categorical   & situation that the shot is being a goal (0: no goal, 1: goal)\\
        \texttt{minute}            & continuous    & minute of shot between 1 and 90 + possible extra time \\
        \texttt{home and away}     & categorical   & status of the shooting team (home or away)\\
        \texttt{situation}         & categorical   & situation at the time of the event (Direct freekick, From corner, Open play, Penalty, Set play) \\
        \texttt{shot type}         & categorical   & type based on the limb used by the player to shot (Head, Left foot, Right foot, Other part of the body) \\
        \texttt{last action}       & categorical    & last action before the shot (Pass, Cross, Rebound, Head Pass, and 35 more levels) \\
        \texttt{distance to goal}  & continuous    & distance from where the shot was taken to the goal line $([0.295, 84.892]$ in meters)\\
        \texttt{angle to goal}     & continuous    & angle of the throw to the goal line $([0.10^{\circ}, 90^{\circ}])$ \\\bottomrule
    \end{tabular}
    \label{tab:features}
    \end{center}
\end{table}\vspace{0.3cm}

The summary statistics of the shots and goals, such as the number ($\#$) of matches, shots, goals, the mean ($\mu$) of shots and goals per match, and the conversion percent ($\%$) of a shot to goal, per league over seven seasons are given in Table \ref{tab:summary}. 

\begin{table}[h]
    \centering
    \caption{The summary statistics of the shots and goals, such as the number ($\#$) of matches, shots, goals, the mean ($\mu$) of shots and goals per match and the conversion percent ($\%$) of a shot to goal for per league over seven seasons}
    \label{tab:summary}
    \begin{tabular}{lcccccc}\toprule
        League     &$\#$Match &$\#$Shot &$\mu_{Shot}$& $\#$Goal& $\mu_{Goal}$& $\%$ \\\toprule
        Bundesliga & 2,141    & 55,129 & 25.7    &6,161   & 2.88    & 11.2 \\
        EPL        & 2,650    & 66,605 & 25.1    &6,951   & 2.62    & 10.4 \\
        La Liga    & 2,648    & 62,028 & 23.4    &6,854   & 2.59    & 11.0 \\
        Ligue 1    & 2,557    & 61,053 & 23.9    &6,438   & 2.52    & 10.5 \\
        Serie A    & 2,659    & 70,615 & 26.6    &7,252   & 2.73    & 10.3\\\midrule
        Mean       & 2,531    & 63,086 & 24.9    &6,371   & 2.67    & 10.7\\\midrule
        Total      & 12,655   & 315,430& -       &33,656  & -       & - \\\bottomrule
    \end{tabular}
\end{table}\vspace{0.3cm}

According to the summary statistics, the conversion percent of all leagues is 0.107 and the Bundesliga has the highest rate is 0.112 while the Serie A has the lowest rate of 0.103. The interesting statistics related to Serie A is the percent of conversion to goals is the lowest, however it is the league with the highest number of shots per match.

\begin{figure}[h]
    \centering
    \caption{The distribution of \texttt{angle to goal} and \texttt{distance to goal} of shots regarding goal status in the last seven seasons of top-five European football leagues}
    \vspace{3mm}
    \label{fig:shot_distribution}
    \includegraphics[scale = 0.32]{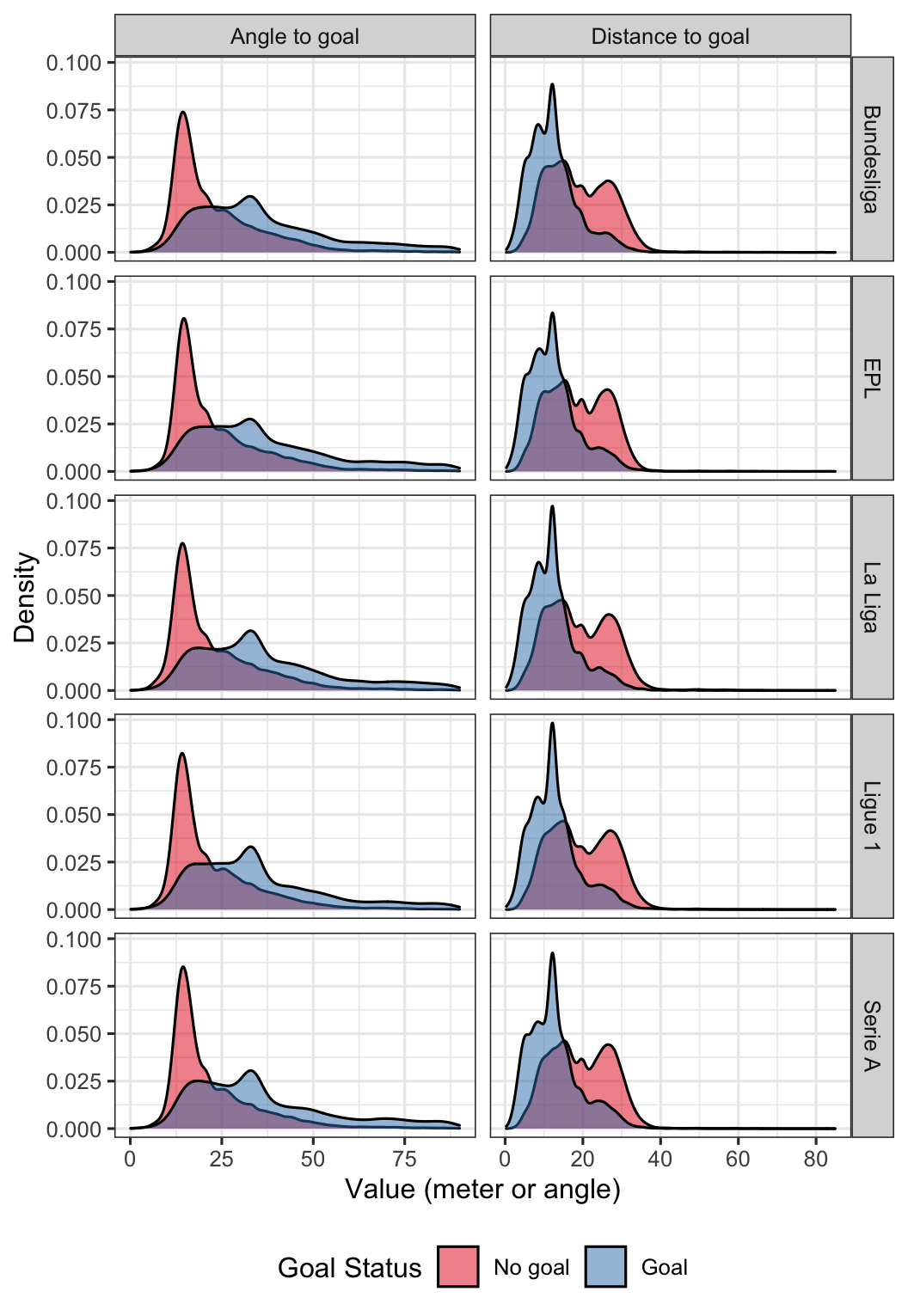}
\end{figure}

Similar to the results of Robberechts and Davis \cite{Rob_and_Davis}, the similarity of the distributions of the  \texttt{distance to goal} and \texttt{angle to goal} from different leagues can be seen in Fig. 1. We want to explore the similarity in terms of those features, since they are the main two variables in the xG models. Fig. 1 shows that: (1) the distribution of the \texttt{distance to goal} of shots seems similar for each league, (2) the range of \texttt{distance to goal} is between 0-25 meters, (3) the optimal angle to goal is about $30^{\circ}$. The distribution of \texttt{angle to goal} and \texttt{distance to goal} of shots regarding goal status in the last seven seasons of the leagues seems similar. It is observed that the distributions of the two most important features affecting the probability of goal between leagues are similar.

\subsection{Pre-processing of the Data}

\begin{figure}[h]
    \centering
    \caption{The standard dimension of a football pitch}
    \label{fig:football_pitch}
    \includegraphics[trim={0 18cm 0 13cm}, clip, scale = 0.2]{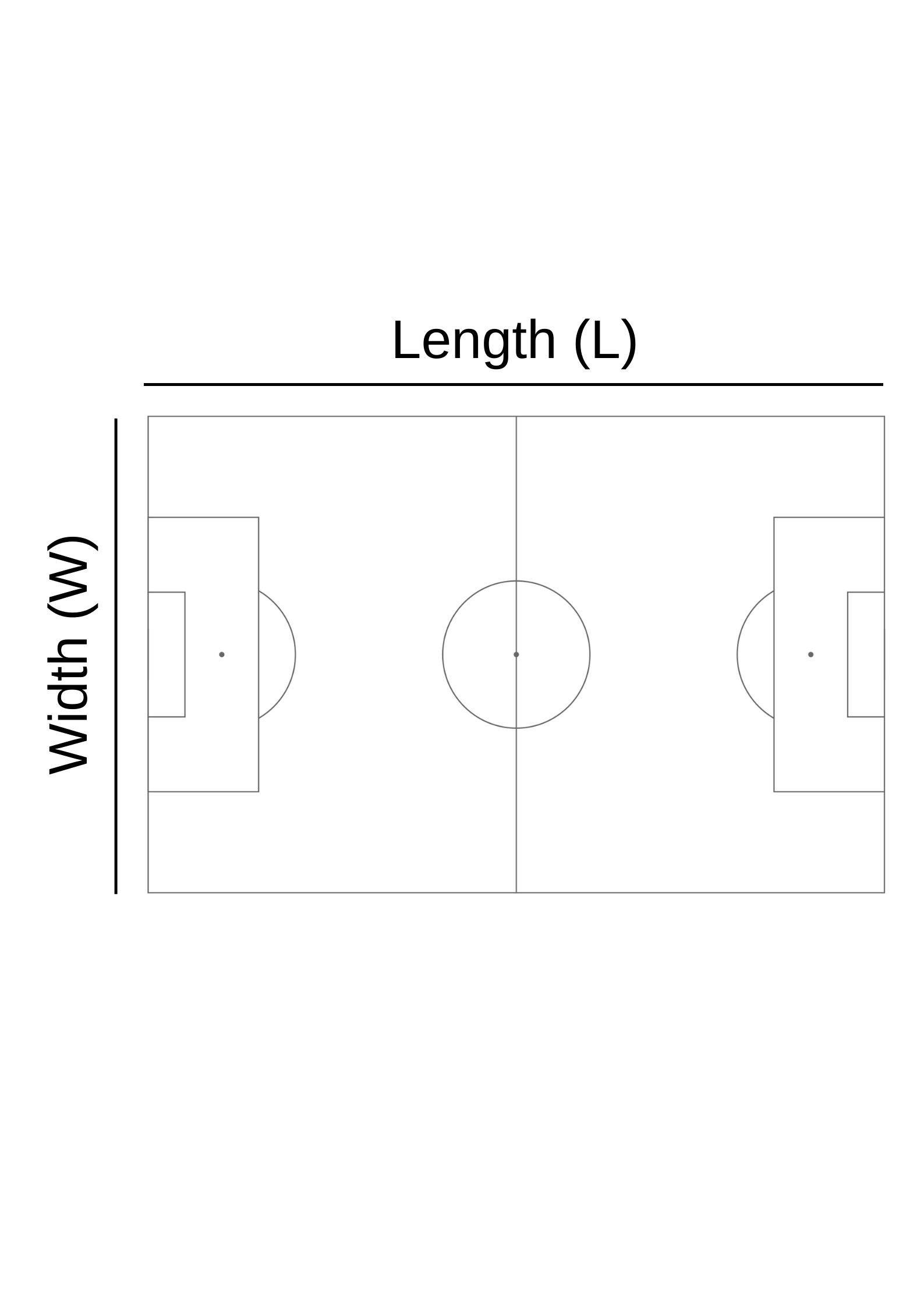}
\end{figure}

The pre-processing steps are necessary before modeling such as the transformation of some features. The location of shots is given in the coordinates system in the dataset as $L_i \in [0, 1]$ and $W_i \in [0, 1]$ as in Fig \ref{fig:football_pitch}. We must calculate the distance and angle to goal of the shots which are the two most important features in xG models based on the coordinate values because the coordinates are not meaningful in the interpretation of the model. Before calculating these features, we standardized a football pitch is $L = 105$m $\times$ $W = 68$m in size, however some pitches may have different dimensions in reality. The size of the pitch is that the length should be between 90 and 120 meters and the width should be between 45 and 90 meters are limited by the rules of The International Football Association Board\footnote{\href{https://digitalhub.fifa.com/m/5371a6dcc42fbb44/original/d6g1medsi8jrrd3e4imp-pdf.pdf}{https://digitalhub.fifa.com/m/5371a6dcc42fbb44/original/d6g1medsi8jrrd3e4imp-pdf.pdf}}.

\noindent The following transformation are used to calculate the distance ($X^{DTG}$) and angle to goal ($X^{ATG}$) features:

\begin{equation}
    X^{DTG}_i = \sqrt{[105 - (L_i \times 105)] ^ 2 + [34 - (W_i \times 68) ^ 2]}
\end{equation}

\noindent where $L$ and $W$ are the coordinates of a shot.

\begin{equation} 
    X^{ATG}_i = \left|\frac{a_i}{b_i} \times \frac{180}{\pi}\right|
\end{equation}

\noindent where $a_i = \arctan[7.32 \times [105 - (L_i \times 105)]]$ and $b_i = [105 - (L_i \times 105)]^2 + [34 - (W_i \times 68)] ^ 2 - (7.32 / 2) ^ 2$. We used $X^{DTG}$ and $X^{ATG}$ features in model training instead of the original coordinates $L_i$ and $W_i$ given in the raw data. The following point should be noted: the transformation is implemented on the raw data according to the one-goal, because the raw data was aggregated for one-goal.

\subsection{Model Training}

We use the forester \cite{forester} AutoML tool to train various tree-based classification models from XGBoost \cite{xgboost}, randomForest \cite{breiman}, LightGBM \cite{lightgbm}, and CatBoost \cite{catboost} libraries. These models do not provide any pre-processing steps like missing data imputation, encoding, or transformation and show quite good performance in the presence of outlier(s) in the dataset which is used to train models. We use the train-test split (80-20) to train and validate the models. Moreover, another advantage of the forester is that provides an easy connection to DALEX \cite{dalex} model explanation and exploration ecosystem.

\subsection{Data balancing}
\label{sec:balancing}

Imbalancedness is a kind of problem when one of the classes of the target variable is rare over the sample in the classification task. In this case, the model learns more from the majority class that which lead to poor classification performance of the minority class. The imbalance problem between the class of target feature is considered a separate field called imbalanced learning in machine learning. In imbalanced learning, there are three main strategies to train the models \cite{guo}: (1) balancing the dataset, by using over or under-sampling methods (2) using cost-sensitive learners, and (3) using ensemble learning models. 

The target feature we used in the model is imbalanced (90$\%$-10$\%$), and to handle this problem we prefer to use a balancing strategy which is the random over-sampling method provided by the ROSE package in R \cite{rose}. It consists of a smoothed bootstrap-based technique which is proposed by Menardi and Torelli \cite{menardi}. 

\section{Exploration and explanation of the proposed xG model}

\vspace{3mm}

In this section, the performance of the trained xG models are investigated in terms of several metrics under different sampling strategies. Then, the best performing xG model is compared with the alternatives in the literature. The aggregated profiles that used to analysis model behavior of the proposed xG model is introduced for evaluating the performance at player and team levels in the last part. Moreover, the practical applications of aggregated profiles are given on the xG model.

\subsection{Model Performance}

For the model level exploration, the first step is usually related to model performance. Different measures may be used such as precision, recall, F1, accuracy, and AUC. However, these measures do not measure the performance of a classification model in case of imbalanced data. The Mathews correlation coefficient, brier score, log-loss and balanced accuracy are used to measure the classification ability of both of the classes in this task. Unfortunately, there are only a few papers have used appropriate measures regarding this problem in the literature related to the xG model. Considering this situation, we reported the results in terms of both of these two groups of measures. During the study of this paper, we restricted the analysis to a comparison of model performance between train and test data. The performance measures of the random forest, catboost, lightgbm, and xgboost models trained on the train, over-sampled train, and under-sampled train set are calculated and given in Table \ref{tab:performance}.

\begin{table}[h]
    \centering
    \small
    \caption{Performance of trained xG models}
    \label{tab:performance}
    \begin{tabular}{p{1.6cm}p{1cm}p{1cm}p{1cm}p{1cm}p{1cm}p{1cm}p{1cm}p{1cm}p{1cm}p{1.2cm}}\toprule
        Model & Sampling & Recall & Precision & F1 & Accuracy & AUC & MCC & Brier Score & Log-loss & Balanced Accuracy \\\toprule
        \multirow{3}{*}{random forest} & over &  \textbf{0.958} & \textbf{0.922} & \textbf{0.940} & \textbf{0.939} & \textbf{0.985} & \textbf{0.879} & 0.071 & 0.270 & \textbf{0.939}  \\
    & under    & 0.858 & 0.882 & 0.870 & 0.871 & 0.954 & 0.743 & 0.104 & 0.352 & 0.871 \\
    & original & 0.304 & 0.888 & 0.453 & 0.921 & 0.975 & 0.493 & \textbf{0.051} & \textbf{0.173} & 0.649 \\\midrule
        \multirow{3}{*}{catboost} & over & 0.740 & 0.762 & 0.751 & 0.755 & 0.839 & 0.510 & 0.164 & 0.495 & 0.755\\
    & under    & 0.728 & 0.756 & 0.742 & 0.745 & 0.828 & 0.492 & 0.169 & 0.507 & 0.746 \\
    & original & 0.198 & 0.722 & 0.311 & 0.906 & 0.823 & 0.347 & 0.074 & 0.261 & 0.594 \\\midrule
        \multirow{3}{*}{xgboost} & over & 0.727 & 0.749 & 0.738 & 0.742 & 0.821 & 0.484 & 0.172 & 0.517 & 0.742\\
    & under    & 0.727 & 0.757 & 0.742 & 0.746 & 0.823 & 0.492 & 0.171 & 0.513 & 0.746  \\
    & original & 0.185 & 0.721 & 0.294 & 0.905 & 0.819 & 0.334 & 0.075 & 0.263 & 0.588 \\\midrule
        \multirow{3}{*}{lightgbm} & over & 0.721 & 0.748 & 0.734 & 0.739 & 0.818 & 0.480 & 0.173 & 0.520 & 0.739\\
    & under    & 0.719 & 0.753 & 0.736 & 0.741 & 0.820 & 0.482 & 0.172 & 0.518 & 0.741 \\
    & original & 0.183 & 0.708 & 0.291 & 0.904 & 0.817 & 0.328 & 0.075 & 0.264 & 0.587 \\\bottomrule
    \multicolumn{11}{l}{*The best value is given in bold for each metric.}
    \end{tabular}
\end{table}\vspace{0.3cm}

\subsection{Comparison of Model Performance}

We compared the performance of our proposed models with the models in the literature \cite{Eggels}, \cite{pardo}, \cite{tippana}, \cite{Anzer_and_Bauer}, \cite{haaren}, \cite{umami}, \cite{fernandez} in terms of precision, recall, accuracy, F1, AUC, log-loss, Brier score, and mean absolute error (MAE) in Table \ref{tab:comparison}. We decided to use these measures because the authors of these papers reported the performance of the models. The reason for the empty cells in the table is that these values for the relevant models have not been reported in these papers. 

\begin{table}[h]
    \small
    \caption{Performance Comparison of the proposed xG model with the models in the literature}
    \begin{center}
    \begin{tabular}{p{3.5cm}p{2.7cm}p{0.9cm}p{0.9cm}p{0.9cm}p{0.9cm}p{0.9cm}p{0.9cm}p{0.6cm}}\toprule
        Paper & Model & Precision & Recall &F1 & AUC & Log-loss & Brier score & MAE \\\toprule
        \multirow{4}{*}{Eggels et al. (2016)} & random forest & 0.785 & 0.822 & 0.800 & 0.814 & - & - & - \\
        & decision tree & 0.698 &  0.678 & 0.676 & 0.677 & - & - & - \\
        & logistic regression & 0.715 & 0.650 & 0.673 & 0.697 & - & - & - \\
        & ada-boost & 0.624 &  0.773 & 0.688 & 0.670 & - & - & -\\\midrule
        
        \multirow{3}{*}{Pardo (2020)} & logistic regression &  - & - & - &  - & 0.261 & - & - \\
        & xgboost &  - & - & - & - & 0.257 & - & - \\
        & neural network &  - & - & - & - & 0.260 & - & - \\\midrule
        
        Tippana (2020) & Poisson regression &  - & - & - & - & - & - & 6.5 \\\midrule
        
        \multirow{4}{*}{Anzer and Bauer (2021)} & gbm &  0.646& 0.181 & - & 0.822 & - & - & - \\
        & logistic regression &  0.611 & 0.108 & - & 0.807 & - & - & -\\
        & ada-boost &  0.548 & 0.201 & - & 0.816 & - & - &  -  \\
        & random forest &  0.611 & 0.163 & - & 0.794 & - & - & - \\\midrule
        
        Haaren (2021) & boosting machine &  - & - & - & 0.793 & - & 0.082 & - \\\midrule
        
        Umami et al. (2021) & logistic regression &  - & \textbf{0.967} & - & - & - & - & - \\\midrule
        
        Fernandez et al. (2021) & xgboost &  - & - & - & - & \textbf{0.254} & - & -\\\midrule
        
        \multirow{4}{*}{Our models} & random forest & \textbf{0.922} & 0.958 & \textbf{0.940} & \textbf{0.985} & 0.270 & \textbf{0.071} & \textbf{2.0} \\
        & catboost  & 0.762 & 0.740 & 0.751 & 0.839 & 0.495 & 0.164 & 2.9\\
        & xgboost   & 0.749 & 0.727 & 0.738 & 0.821 & 0.517 & 0.172 & 3.1\\
        & lightgbm  & 0.748 & 0.721 & 0.734 & 0.818 & 0.520 & 0.173 & 3.0 \\\bottomrule
        \multicolumn{9}{l}{*The best value is given in bold for each metric.}
    \end{tabular}
    \label{tab:comparison}
    \end{center}
\end{table}\vspace{0.3cm}

Eggels et al. \cite{Eggels}, Pardo \cite{pardo}, and Anzer and Bauer \cite{Anzer_and_Bauer} trained several xG models and reported their performances. The random forest, xgboost, and gbm models outperform others, respectively in these papers. It is seen that our proposed random forests model outperforms the others in terms of precision, F1, AUC, Brier score, and MAE. The model proposed by Fernandez et al. \cite{fernandez} has a lower log-loss value than the model that we proposed. However, no information is reported about the performance of their model in terms of recall and precision, so it is difficult to be certain of its performance. It is same for the model proposed by Umami et al. \cite{umami}. 

\subsection{Model Behavior: The Aggregated Profiles}

The XAI tools are classified under two main sections are local and global levels. The local-level explanations are used to explain the behavior of a black-box model for a single observation while the global-level explanations are used for an entire dataset. However, our need is to explain the model for a group of observations, i.e. for a player or team. That's why we introduce the aggregated profiles (AP) which can be used for a group of observations. 

The idea behind AP is the aggregation of the CP profiles that show how the change of a model’s prediction would change regarding the value of a feature. In other words, the CP profile is a function that describes the dependence of the conditional expected value of the response on the value z of the $j^{th}$ feature \cite{ema}. The AP can be defined simply as the averaging of the CP profiles which are considered. The value of an AP for model $f(.)$ and feature $X_j$ at the point $z$ is defined as follows:

\begin{equation}
    g_{AP}^j (z) = E_\textbf{X}^{-j} [f(\textbf{X}^{j|z})]
\end{equation}

\noindent where $g_{AP}$ is the expected value of the model predictions when $X_j$ is fixed at $z$ over the marginal distribution of $\textbf{X}_{j|z}$. The distribution of $\textbf{X}_{j|z}$ can be estimated by using the mean of CP profiles for $X_j$ as an estimator of the AP:

\begin{equation}
    \hat{g}^j_{AP} (z) = \frac{1}{k} \sum_{i = 1}^{k} f(\textbf{x}^{ij|z})
\end{equation}

\noindent where $k$ is the number of profiles that are aggregated. The difference between the AP and PDP is the number of aggregated profiles. The PDP is the aggregation of all profiles which are calculated on the entire dataset while the AP is the aggregation of several profiles. The innovative part of this paper is to propose a performance evaluation by aggregating CP profiles of shots (i.e. observations) taken by a team or player during the period of interest. This evaluation can be defined as post-game analysis if the period is a game or games. Also, it can be used after the first half of a match to decide the second-half strategy. In this way, answers to what-if questions can be created based on teams or players, not based on shots that are not meant to evaluate.

\subsection{Practical Applications of the AP in Football}

In this part, we demonstrate the practical applications of the aggregated profiles, which is constructed on the proposed xG model, for team and player levels in football. Firstly, consider the match of Schalke 04 vs. Bayern Munich which is played in Bundesliga on Jan 24, 2021. The end-of-match statistics such as number of goals ($\#$Goal), expected goal (xG), number of shots ($\#$Shot), mean angle to goal ($\mu_{ATG}$), and mean distance to goal ($\mu_{DTG}$) for each team over the match are given in Table \ref{tab:schalke}. 

\begin{table}[h]
    \centering
    \caption{The End-of-match statistics of Schalke 04 vs. Bayern Munich in the match is played on Jan 24, 2021}
    \label{tab:schalke}
    \begin{tabular}{lccccc}\toprule
        Team          & $\#$Goal & xG   & $\#$Shot & $\mu_{ATG}$ & $\mu_{DTG}$ \\\toprule
        Schalke 04    & 0        & 2.67 & 13       & 25.23       & 17.99  \\
        Bayern Munich & 4        & 9.59 & 31       & 27.79       & 16.96  \\\bottomrule
    \end{tabular}
\end{table}\vspace{0.3cm}

According to the match statistics, Bayern Munich took 31 shots while Schalke 04 took 13 shots in this match, 4 of them being goals and Bayern won the game 0-4. The xG values of the teams show the created goal chances over the shots. It is expected that the final score of the match is 3-10 in terms of expected goal. However, Schalke 04 could not find any goal while Bayern Munich found four goals. The offensive efficiency (actual goals - expected goal) of both teams are not good, because they found four goals considering 12.26 (2.67 + 9.59) xG. The observed mean angle to goal is $25.23^{\circ}$ for Schalke 04 and $27.79^{\circ}$ for Bayern Munich, and the observed mean distance to goal is about $18$ meters for Schalke 04 and $17$ meters for Bayern Munich. 

The AP of the Schalke 04 and Bayern Munich for angle and \texttt{distance to goal} are given in Fig. 3. In the figure of AP, X-axis represents the value of the interested feature, and y-axis represents the average prediction of the xG for per-shot. The vertical dotted lines show the mean observed value of the feature per team in the match.

\begin{figure}[h]
    \centering
    \caption{The aggregated xG profiles of Schalke 04 and Bayern Munich for angle and distance to goal in the match is played on Jan 24, 2021}
    \label{fig:app}
    
    \vspace{3mm}
    \centering
    \includegraphics[trim = 0cm 0cm 0cm 4cm, clip = true,
    scale = 0.1, width=0.55\textwidth]{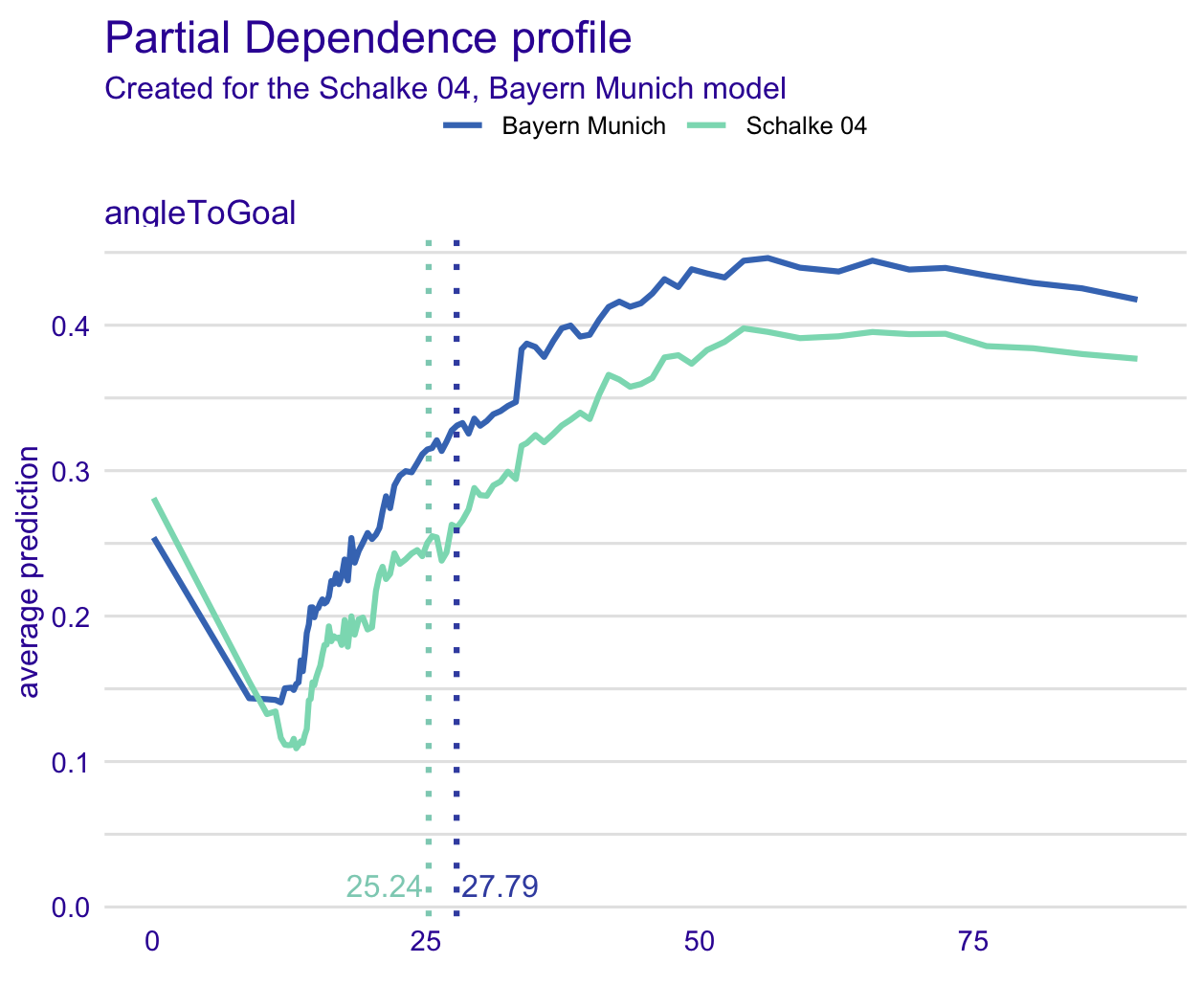}

    \centering
    \includegraphics[trim = 0cm 0cm 0cm 6cm, clip = true,
    scale = 0.1, width=0.55\textwidth]{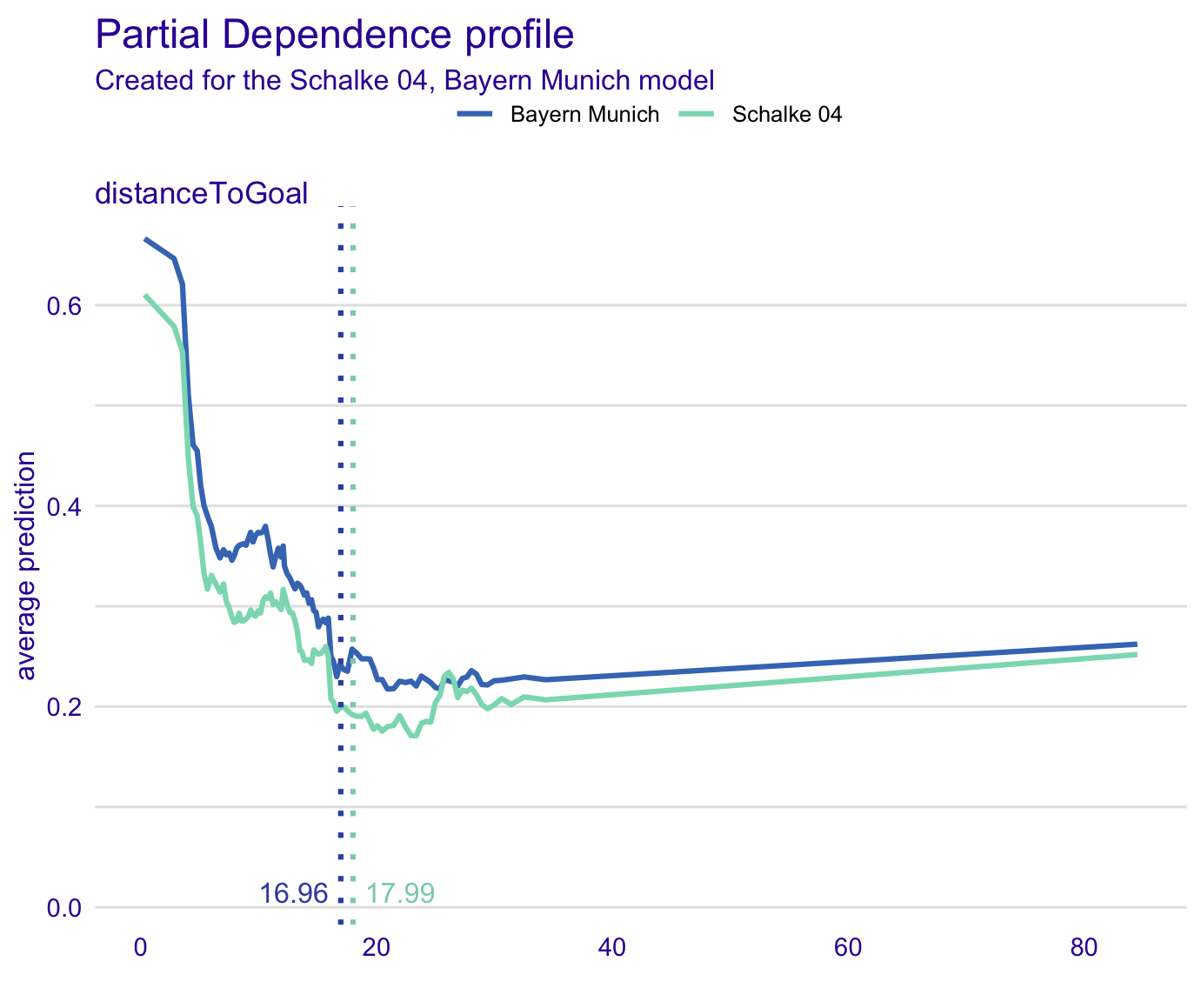}
    
\end{figure}

\noindent As known, if the shots are taken at a steeper angle and closer to the goal, the value of xG increases. It is seen that the average of the xG is about constant after 30 meters according to the AP for distance to goal. Evaluating the average distances to goal, Schalke 04 is farther from the goal than Bayern. This may be one of the reasons why they did not find any goal. They could have increased the average xG for pre-shot by around 40 percent if they reduced the average distance to goal from 18 meters to 15 meters. Moreover, if Schalke had taken the shots with an average angle of $35^{\circ}$ instead of an average angle of $25^{\circ}$, the average value of xG for per-shot would have increased by $20\%$. This potential change can be considered to improve the team's performance for the next match(s). As seen, AP provides to evaluate team performance after a match and determine how the team can increase the value of xG with possible improvements during the game.

Secondly, consider the player performance of Burak Yilmaz is the striker of Lille OSC from Ligue 1, Lionel Messi is the midfielder of FC Barcelona from La Liga, and Robert Lewandowski is the striker of Bayern Munich from Bundesliga in the season of 2020-21. According to the end-of-season statistics given in Table~\ref{tab:lewa}, they took 66, 195, 132 shots, and 16, 30, 40 of them being goals during the season, respectively. The player with the highest ability to convert shots into goals is Robert Lewandowski with $30\%$, Burak Yilmaz is second with $25\%$ and Lionel Messi is third with $15\%$. The ability isn't just about skill, it may also related to how players defend against and the average angle and distance from which they use the shots. Robert Lewandowski shot from the shorter distance and more right angle on average, while the other players shot from the relatively long distance and narrow angle. It can be roughly said that the percent of the players converting shots into goals, in other words the created xG values, are correlated with distance and angle to goal. 

\begin{table}[h]
    \centering
    \caption{The End-of-season statistics of Burak Yilmaz (BY), Lionel Messi (LM), and Robert Lewandowski (RL) in the season of 2020-21}
    \label{tab:lewa}
    \begin{tabular}{lcccccc}\toprule
        Player  &$\#$Game& $\#$Goal & xG    & $\#$Shots & $\mu_{ATG}$ & $\mu_{DTG}$ \\\toprule
        BY      &24      & 16       & 24.77 & 66        & 22.33       & 19.43 \\
        LM      &35      & 30       & 70.00 & 195       & 21.66       & 19.23 \\
        RL      &28      & 40       & 65.71 & 132       & 34.69       & 12.94 \\
        \bottomrule
    \end{tabular}
\end{table}\vspace{0.3cm}

\noindent The AP of Burak Yilmaz, Lionel Messi, and Robert Lewandowski for angle and distance to goal in the season of 2020-21 are given in Fig. 4. The vertical dotted lines show the mean observed value of the feature per player in the season.

\begin{figure}[h]
    \centering
    \caption{The aggregated profiles of Burak Yilmaz, Lionel Messi, and Robert Lewandowski for angle and distance to goal in the season of 2020-21}
    \label{fig:lewa}
    
    \vspace{3mm}
    \centering
    \includegraphics[trim = 0cm 0cm 0cm 4cm, clip = true,
    scale = 0.1, width=0.55\textwidth]{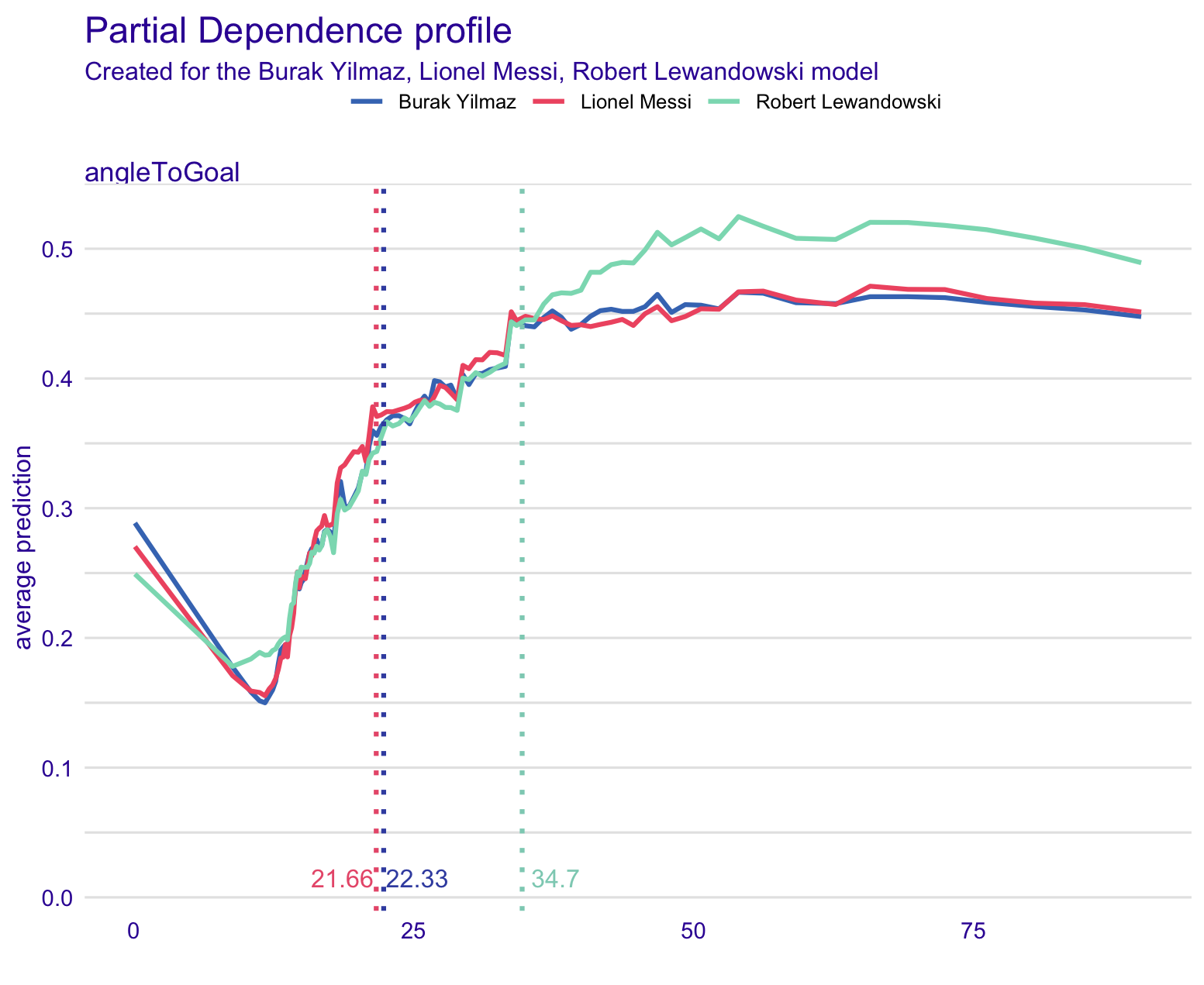}

    \centering
    \includegraphics[trim = 0cm 0cm 0cm 6cm, clip = true,
    scale = 0.1, width=0.55\textwidth]{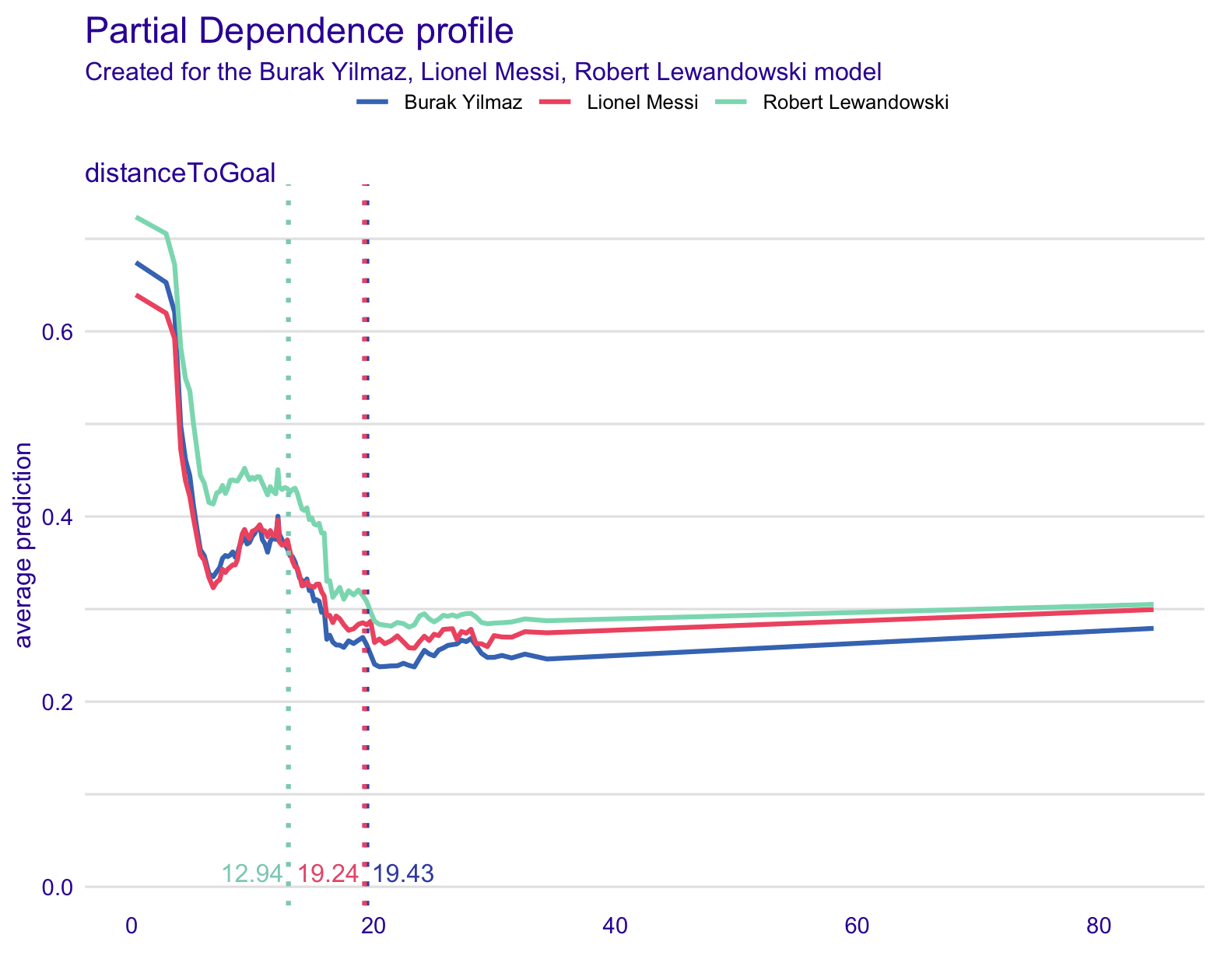}
    
\end{figure}

The average value of the xG is about similar for each player for distance to goal. Only the average value of Robert Lewandowski is slightly higher than the others after 10 meters. It means that if the players try to take shots at 15 meters from the goal instead of the observed mean distances seen in Table \ref{tab:lewa}, the average expected goal of the player per shot may increase about $20\%$. When the AP is examined in terms of the feature is angle to goal, the average xG of the player is about same from the start point to $25^{\circ}$, the AP of Robert Lewandowski is getting increase after this angle. It can be said that if the players try to take shots at $50^{\circ}$, the highest average xG of Robert Lewandowski will be about 0.5. At the player level as well as at the team level, the AP provides pretty practical information about performance evaluation and provisioning, and it is also very useful for comparing the players that play at similar position.
\section{Discussion}

Domain-specific applications of XAI tools enable key insights to be extracted from a black-box model. This paper focuses the practical application of the AP for explaining more than one observation, not an observation or entire dataset in this context. These tools may be referred to as \textit{semi-global explainers} for easier understanding in the XAI domain. It is seen from the examples discussed: AP can be used to extract provisions for performance analysis of a team or a player from xG models, which have been frequently used in football in recent years. In addition, comparisons of similar players and teams can be made in terms of interested features. Since this approach can be used for other sports branches such as ice hockey where the xG models are used, its widespread effect is not be limited to football.

\begin{figure}[h]
    \centering
    \caption{The behavior comparison of the random forest models trained on original, over-sampled, and under-sampled data in terms of PDP curves}
    \label{fig:imbalance}
    
    \vspace{3mm}
    \centering
    \includegraphics[trim = 0cm 0cm 0cm 4cm, clip = true,
    scale = 0.20, width = 0.55\textwidth]{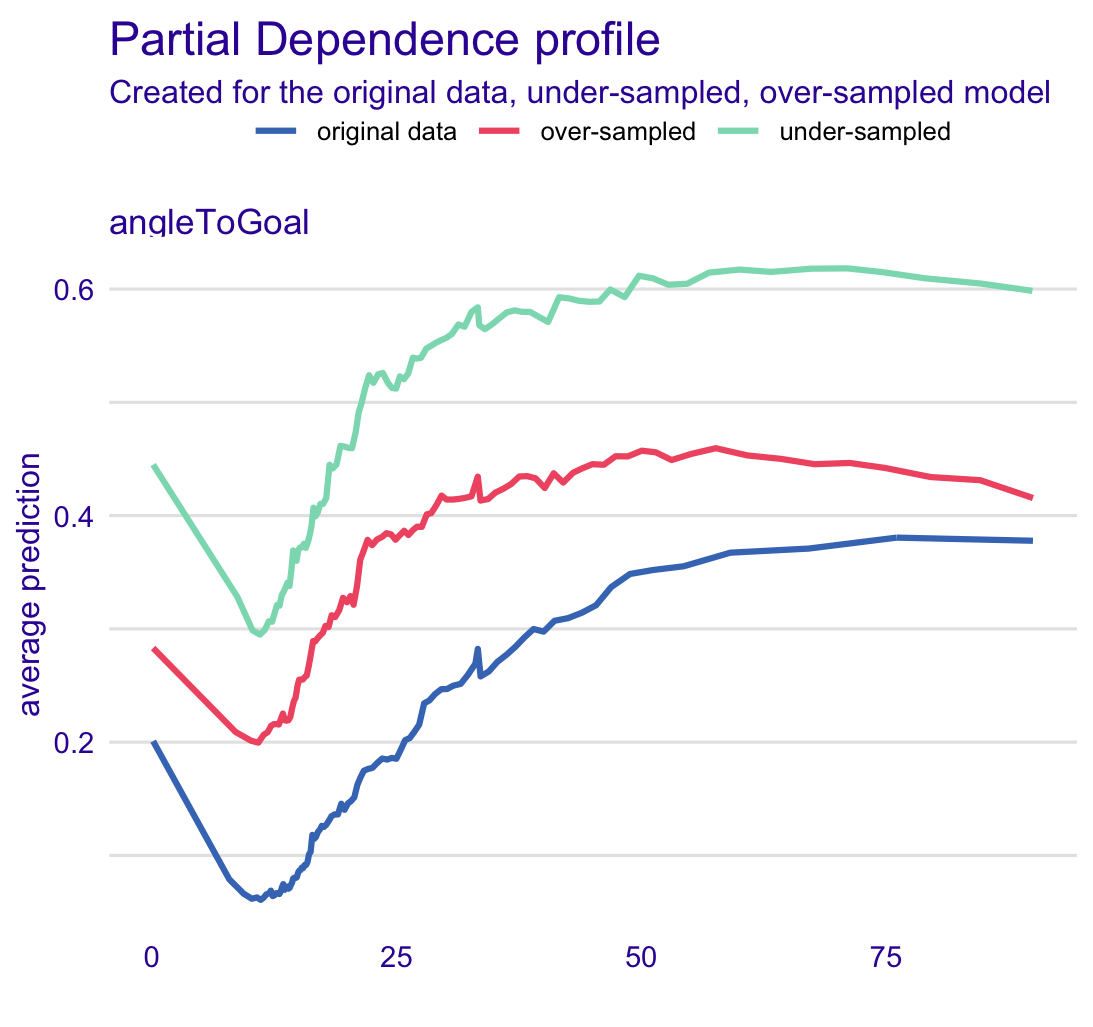}

    \centering
    \includegraphics[trim = 0cm 0cm 0cm 6cm, clip = true,
    scale = 0.20, width = 0.55\textwidth]{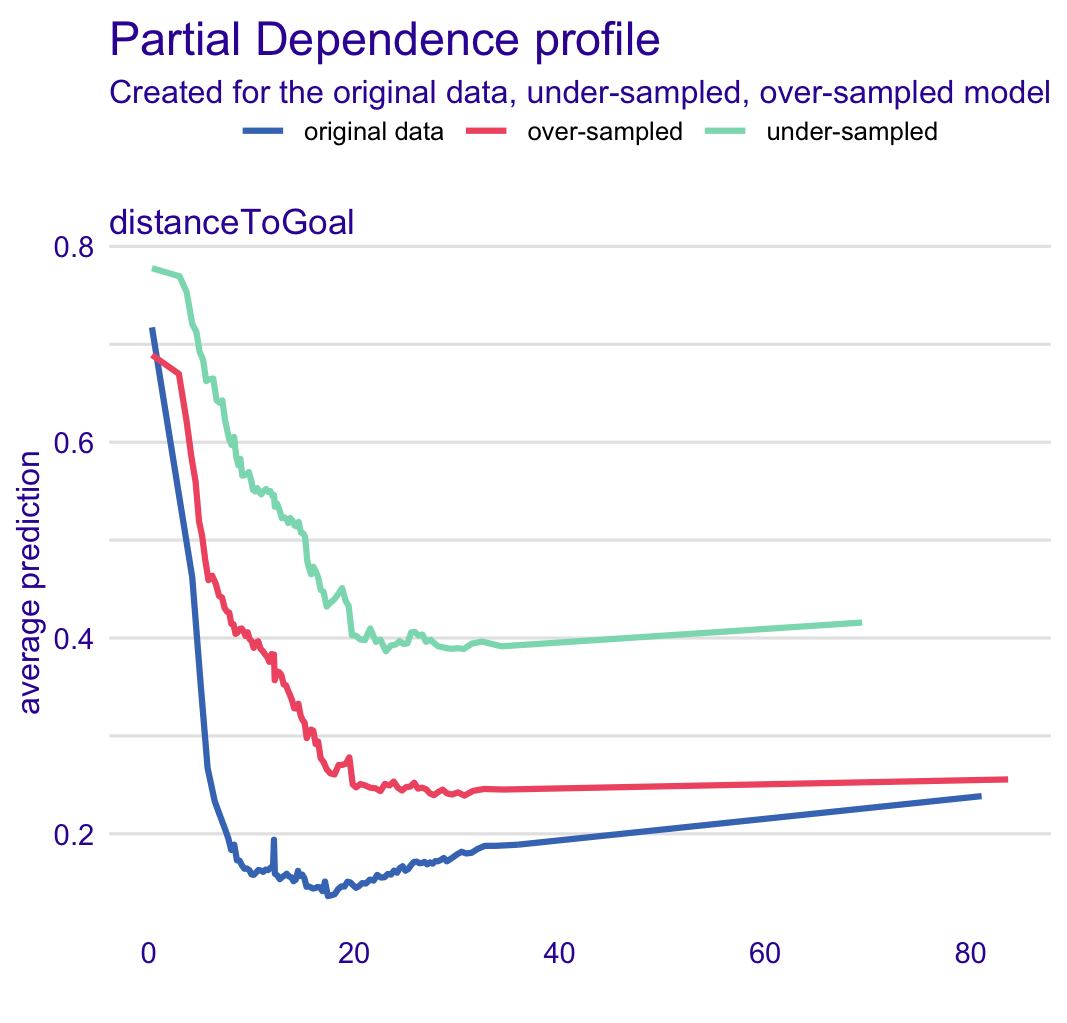}
    
    \centering
    \includegraphics[trim = 0cm 0cm 0cm 6cm,
    clip = true,
    scale = 0.20, width = 0.55\textwidth]{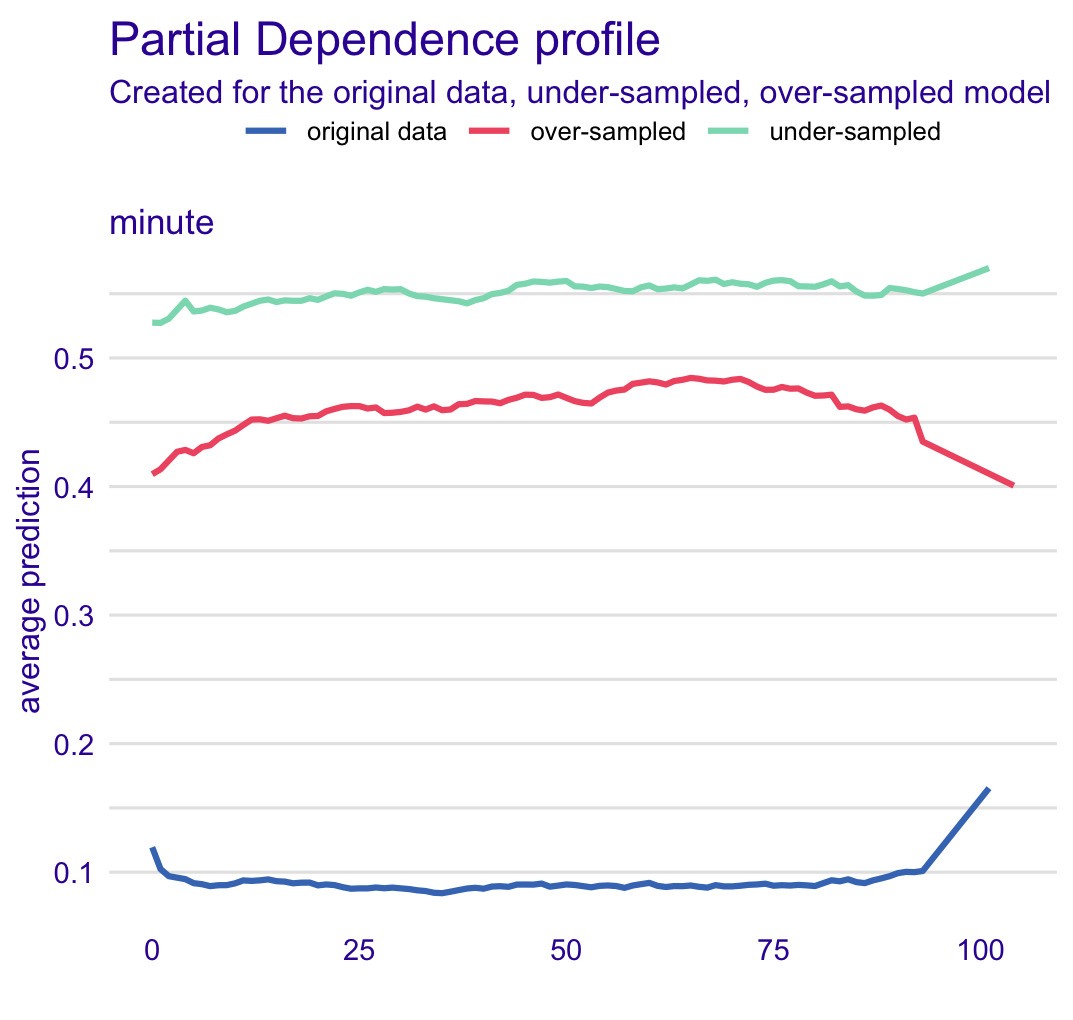}
    
\end{figure}

Another discussion we want to mention in this paper is the effect of balancing methods on the model's behavior. It is known that these methods for imbalanced datasets in binary classification tasks is a very commonly used solution to improve the prediction performance of ML models in both the classes. However, how balancing the observation of classes causes effects on the model behavior is a subject that has not been discussed yet. The only paper in the literature about this is Patil et al. \cite{patil} discussed that whether the model is reliable or not after balancing. They decided to verify the reliability of the models and oversampling method with the help of the feature importance which is one of the XAI techniques. Their results demonstrate that the higher accuracy obtained by the over-sampled dataset while ensuring that the oversampling does not alter the feature correlation of the original dataset. This paper does not satisfy to decide the change of model behavior, because it only examined the change of order of features' importance in the model. The point we want to raise is the model behavior in terms of PDP's values, because it provides more detailed information than the feature importance to detecting the change in the model behavior. Thus, we compared the behavior of the original, over-sampled, and under-sampled versions of our proposed model using PDP curves for some features in Fig \ref{fig:imbalance}. 

It is seen that behavior of the PDP curves for the feature is distance to goal are not same for the models. For the model trained on original data, the PDP curve is increasing from start point to 20 meters, then slowly increasing after some fluctations between 20 and 30 meters. However, the fluctations are seen on different range of the feature for the models trained on over and under-sampled data. This is a sign that there has been a change in model behavior after using balancing methods. We would like to draw attention to the need for careful consideration of this situation, and start a discussion on this subject in the literature based on our findings.
\section{Conclusion}

The papers to date aim that to evaluate a team or player's performance that only consider the output of the xG model. Comparing the actual goals and expected goals, they provide some statistics based on these difference for evaluating the defensive and offensive performance of a team or a player. However, we focused to use the xG model's behavior to observe the relationship between features and response which is the xG value. In this way, we can suggest to how a team's performance can be improved by changing the strategies based on the features that effect the xG value such as distance to goal, angle to goal, and others. To do this, we first proposed an accurate xG model which is trained a random forests model on the data consist seven seasons of top-five European leagues. This model predicts both possible outcomes of the output of the xG model, goal and no goal, better than the alternatives in the literature. To obtain this model, we balanced the data by using the random over-sampling method to solve the imbalance problem that is often ignored in similar papers. Thus, the model we proposed learned quite well from both of the classes. The interesting thing in this process is that we detected some changes in the behavior of the model trained on the data obtained after over-sampling through PDP curves. We have included a detailed discussion on this problem in Discussion section.

We evaluated the performance at the level of the team and the player by using the accurate xG model we proposed, in the practical application part. The AP of the model on the team level show to relationship between the interested features and the target variable is the average of xG values predicted over the observations. The usage of AP is practical to see the effect of the changes on the values of interested variable on the xG value, and also comparison of the players who are play in similar positions. For example, it can be extracted from Fig. 4 that the average xG of Robert Lewandowski may be higher than other players if he shoots from a steeper angle. As seen, detailed information about performance evaluation can be obtained with the practical use of AP, which is one of the XAI tools, on the xG models. This is a good example of how using XAI tools contributes in different application areas.

\section*{Acknowledgement}

The work on this paper is financially supported by the NCN Sonata Bis-9 grant 2019/34/E/ST6/00052.

\section*{Supplemental Materials}

The R codes needed to reproduce the results can be found in the following GitHub repository: \href{https://github.com/mcavs/Explainable_xG_model_paper}{https://github.com/mcavs/Explainable\_xG\_model\_paper}.

\bibliographystyle{unsrtnat}


\begin{thebibliography}{1}



\bibitem{green} 
S. Green.
\newblock Assessing the performance of premier league goalscorer.
\newblock {\em OptaPro Blog}, 2012.

\bibitem{cardoso} 
F. Cardoso, S. González-Víllora, J. Guilherme, and I. Teoldo. 
\newblock Young Soccer Players With Higher Tactical Knowledge Display Lower Cognitive Effort.
\newblock {\em Percept Mot Skills}, vol. 126, pp. 499--514, 2019.

\bibitem{rathke} 
A. Rathke. 
\newblock An examination of expected goals and shot efficiency in soccer.
\newblock {\em J Hum Sport Exerc, vol. 12, pp. 514-–529, 2017}.

\bibitem{tippana} 
T. Tippana.
\newblock How accurately does the expected goals model reflect goalscoring and success in football?, 
\newblock {\em Bachelor's Thesis}, Aalto University, 2020.

\bibitem{pardo} 
M. Pardo.
\newblock Creating a model for expected goals in football using qualitative player information. 
\newblock {\em Master’s thesis}, Universitat Politecnica de Catalunya, 2020. 

\bibitem{herbinet} 
C. Herbinet.
\newblock Predicting football results using machine learning techniques. 
\newblock {\em MEng thesis}, Imperial College London, 2018.

\bibitem{wheatcroft} 
E. Wheatcroft and E. Sienkiewicz.
\newblock A probabilistic model for predicting shot success in football. 
\newblock {\em arXiv preprint}, arXiv:2101.02104, 2021.

\bibitem{bransen} 
L. Bransen and J. Davis.
\newblock Women’s football analyzed: interpretable expected goals models for women.
\newblock {\em In AI for Sports Analytics (AISA) Workshop at IJCAI 2021}. 

\bibitem{sarkar_and_kamath} 
S. Sarkar and S. Kamath.
\newblock Does luck play a role in the determination of the rank
positions in football leagues? A study of Europe’s big five. 
\newblock {\em Ann Oper Res}, 2021.

\bibitem{Eggels} 
H. Eggels, R. Van Elk, and M. Pechenizkiy. 
\newblock Explaining soccer match outcomes with goal scoring opportunities predictive analytics. 
\newblock {\em In proceedings of the Workshop on Machine Learning and Data Mining for Sports Analytics 2016 co-located with the 2016 European Conference on Machine Learning and Principles and Practice of Knowledge Discovery in Databases}, 2016.

\bibitem{Anzer_and_Bauer} 
G. Anzer and P. Bauer.
\newblock A goal scoring probability model for shots based on synchronized positional and event data in football (soccer). 
\newblock {\em Front. Sports Act. Living}, vol. 3, pp. 1--15, 2021.

\bibitem{kharrat} 
T. Kharrat, I. G. MacHale, and J. L. Pena.
\newblock Plus-minus player ratings for soccer. 
\newblock {\em European Journal of Operational Research}, vol. 283, pp. 726--736, 2020.

\bibitem{spearman}
W. Spearman.
\newblock Beyond expected goals.
\newblock {\em MIT Sloan Sports Analytics Conference}, 2018.

\bibitem{brechot} 
M. Brechot and R. Flepp.
\newblock Dealing with randomness in match outcomes: how to rethink performance evaluation in European club football using expected goals.
\newblock {\em Journal of Sports Economics}, vol. 21, pp. 335--362, 2020.

\bibitem{fairchild} 
A. Fairchild, K. Pelechrinis, and M. Kokkodis, 
\newblock Spatial analysis of shots in MLS: A model for expected goals and fractal dimensionality.
\newblock {\em Journal of Sports Analytics}, vol. 4, pp. 165--174, 2018.

\bibitem{ema} 
P. Biecek, and T. Burzykowski.
\newblock Explanatory Model Analysis.
\newblock {\em Chapman and Hall/CRC}, New York, 2021.

\bibitem{friedman} 
J. H. Friedman. 
\newblock ``Greedy Function Approximation: A Gradient Boosting Machine''.
\newblock {\em Annals of Statistics}, vol. 29, pp. 1189--1232, 2000.

\bibitem{Rob_and_Davis} 
P. Robberechts and J. Davis.
\newblock How Data Availability Affects the Ability to Learn Good xG Models.
\newblock {\em in Brefeld, U., Davis, J., Van Haaren, J., Zimmermann, A. (eds) Machine Learning and Data Mining for Sports Analytics, MLSA 2020, Communications in Computer and Information Science}, vol. 1324, Springer, Cham, 2020.

\bibitem{worldfootballR} 
J. Zivkovic, and T. ElHabr. 
\newblock worldfootballR: Functions to Extract and Clean World Football (Soccer) Data. https://github.com/JaseZiv/worldfootballR, 2022.

\bibitem{forester} 
H. T. Ly, S. Szmajdzinski, and A. Kozak, 
\newblock forester: Automated Machine Learning Model Solver.
https://github.com/ModelOriented/forester, 2022.

\bibitem{xgboost} 
T. Chen and C. Guestrin. 
\newblock XGBoost: A Scalable Tree Boosting System. 
\newblock {\em In Proceedings of the 22nd ACM SIGKDD International Conference on Knowledge Discovery and Data Mining}, pp. 785--794, 2016.

\bibitem{breiman} 
L. Breiman.
\newblock Random forests.  
\newblock {\em Machine Learning}, vol. 45, pp. 5--32, 2001.

\bibitem{lightgbm} 
G. Ke, Q. Meng, T. Finley, T. Wang, W. Chen, W. Ma, Q. Ye, and T. Liu.
\newblock LightGBM: A Highly Efficient Gradient Boosting Decision Tree. 
\newblock {\em Advances in Neural Information Processing Systems}, pp. 3149--3157, 2017.

\bibitem{catboost} 
A. V. Dorogush, V. Ershov, and A. Gulin. 
\newblock CatBoost: gradient boosting with categorical features support.
\newblock {\em Workshop on ML Systems at NIPS 2017}.

\bibitem{dalex} P. Biecek.
\newblock DALEX: Explainers for Complex Predictive Models in R.  
\newblock {\em Journal of Machine Learning Research}, vol. 19, pp. 1--5, 2018.

\bibitem{guo} 
X. Guo, Y. Yin, C. Dong, G. Yang, and G. Zhou. 
\newblock On the class imbalance problem., 
\newblock {\em In ICNC}, pp. 192–-201, 2008.

\bibitem{rose} N. Lunardon, G. Menardi, and N. Torelli.
\newblock ROSE: A Package for Binary Imbalanced Learning.
\newblock {\em The R Journal}, vol. 6, pp. 79--89, 2014.

\bibitem{menardi} G. Menardi and N. Torelli,
\newblock Training and assessing classification rules with imbalanced data. 
\newblock {\em Data Mining and Knowledge Discovery}, vol. 28, pp. 92--122, 2014.

\bibitem{haaren} 
J. V. Haaren.
\newblock Why would I trust your numbers? On the explainability of expected values in soccer.
\newblock {\em arXiv preprint}, arXiv:2105.13778, 2021. 

\bibitem{umami} 
I. Umami, D. H. Gutama, and H. R. Hatta. 
\newblock Implementing the expected goal (xG) model to predict scores in soccer matches. 
\newblock {\em International Journal of Informatics and Information Systems}, vol. 4, pp. 38--54, 2021.

\bibitem{fernandez} 
J. Fernandez, L. Bornn, and D. Cervone.
\newblock A framework for the fne‑grained evaluation
of the instantaneous expected value of soccer possessions. 
\newblock {\em Machine Learning}, vol. 110, pp. 1389-–1427, 2021. 

\bibitem{patil} 
A. Patil, A. Framewala, and F. Kazi.
\newblock Explainability of SMOTE Based Oversampling for Imbalanced Dataset Problems.
\newblock {\em 3rd International Conference on Information and Computer Technologies}, pp. 41--45, 2020.


\end{thebibliography}

\end{document}